\documentclass[10pt,twocolumn,letterpaper]{article}

\usepackage{cvpr}
\usepackage{times}
\usepackage{epsfig}
\usepackage{graphicx}
\usepackage{amsmath}
\usepackage{amssymb}
\usepackage{booktabs}
\usepackage{multirow}
\usepackage{subcaption}
\usepackage[breaklinks=true,bookmarks=false]{hyperref}

\begin{document}

\title{MelanomaNet: Explainable Deep Learning for Skin Lesion Classification}

\author{Sukhrobbek Ilyosbekov\\
Northeastern University\\
{\tt\small ilyosbekov.s@northeastern.edu}
}

\maketitle
\thispagestyle{empty}
\pagestyle{plain}

\begin{abstract}
    Automated skin lesion classification using deep learning has shown remarkable
    accuracy, yet clinical adoption remains limited due to the ``black box'' nature
    of these models. We present MelanomaNet, an explainable deep learning system
    for multi-class skin lesion classification that addresses this gap through four
    complementary interpretability mechanisms. Our approach combines an EfficientNet
    V2 backbone with GradCAM++ attention visualization, automated ABCDE clinical
    criterion extraction, Fast Concept Activation Vectors (FastCAV) for
    concept-based explanations, and Monte Carlo Dropout uncertainty quantification.
    We evaluate our system on the ISIC 2019 dataset containing 25,331 dermoscopic
    images across 9 diagnostic categories. Our model achieves 85.61\% accuracy with
    a weighted F1 score of 0.8564, while providing clinically meaningful explanations
    that align model attention with established dermatological assessment criteria.
    The uncertainty quantification module decomposes prediction confidence into
    epistemic and aleatoric components, enabling automatic flagging of unreliable
    predictions for clinical review. Our results demonstrate that high
    classification performance can be achieved alongside comprehensive
    interpretability, potentially facilitating greater trust and adoption in
    clinical dermatology workflows. The source code is available at
    \url{https://github.com/suxrobgm/explainable-melanoma}
\end{abstract}

\section{Introduction}

Skin cancer represents one of the most prevalent forms of cancer worldwide,
with melanoma being the most lethal variant responsible for the majority of
skin cancer deaths despite comprising only a small fraction of cases
\cite{siegel2023cancer}. Early detection significantly improves patient
outcomes, with 5-year survival rates exceeding 99\% for localized melanoma
but dropping below 30\% for distant metastatic disease. Dermoscopy, a
non-invasive imaging technique that magnifies skin lesions, has become the
standard clinical tool for melanoma screening. However, accurate
interpretation of dermoscopic images requires substantial expertise, and
diagnostic accuracy varies considerably even among trained dermatologists.

Deep learning has emerged as a promising approach for automated dermoscopic
image analysis, with convolutional neural networks achieving diagnostic
accuracy comparable to or exceeding that of expert dermatologists in
controlled studies \cite{esteva2017dermatologist}. Despite these impressive
results, clinical adoption of AI-assisted dermatology tools remains limited.
A primary barrier is the lack of interpretability inherent in deep neural
networks---clinicians are understandably reluctant to trust diagnostic
recommendations from systems that cannot explain their reasoning.

The ABCDE criteria (Asymmetry, Border irregularity, Color variation, Diameter,
Evolution) represent the established clinical framework for melanoma assessment
\cite{abbasi2004early}. These criteria provide an intuitive and teachable
approach that dermatologists use to communicate findings to patients and
colleagues. An AI system that can relate its predictions to these familiar
clinical concepts would be far more useful and trustworthy than one that
simply outputs class probabilities.

In this work, we present MelanomaNet, an explainable deep learning system that
addresses the interpretability gap through multiple complementary mechanisms:

\begin{itemize}
    \item \textbf{Attention Visualization}: GradCAM++ generates heatmaps
          showing which image regions most influenced the prediction.
    \item \textbf{Clinical Criterion Extraction}: Automated analysis of ABCDE
          features with quantitative scores and visualizations.
    \item \textbf{Concept-Based Explanations}: Fast Concept Activation Vectors
          provide human-interpretable concept importance scores.
    \item \textbf{Uncertainty Quantification}: Monte Carlo Dropout decomposes
          uncertainty into epistemic and aleatoric components.
\end{itemize}

Our contributions include: (1) a multi-modal explainability framework that
bridges deep learning predictions with clinical reasoning, (2) alignment
metrics that validate whether model attention corresponds to clinically
relevant features, and (3) comprehensive uncertainty quantification that
flags unreliable predictions for human review.

\section{Related Work}

\subsection{Deep Learning for Skin Lesion Classification}

The application of deep learning to dermoscopic image analysis gained
significant attention following Esteva et al.'s demonstration that a CNN
trained on clinical images could match dermatologist-level performance
\cite{esteva2017dermatologist}. Subsequent work has explored various
architectures and training strategies. The ISIC challenges have driven
progress in this area by providing standardized benchmarks and large-scale
annotated datasets. Recent approaches have employed transfer learning from
ImageNet-pretrained models, with EfficientNet variants achieving strong
results due to their favorable accuracy-efficiency tradeoff
\cite{tan2019efficientnet}. Gessert et al. \cite{gessert2020skin} demonstrated
that ensembles of EfficientNet models with extensive data augmentation could
achieve top performance on ISIC challenges, though their work focused
primarily on classification accuracy rather than interpretability.

\subsection{Explainability in Medical Imaging}

Explainable AI (XAI) has become increasingly important in medical applications
where understanding model reasoning is critical for clinical acceptance.
Gradient-weighted Class Activation Mapping (Grad-CAM) \cite{selvaraju2017grad}
and its extensions provide visual explanations by highlighting image regions
that contribute most to predictions. In dermatology, attention visualization
helps verify that models focus on lesion features rather than artifacts or
irrelevant background regions. Chattopadhay et al. \cite{chattopadhay2018grad}
proposed Grad-CAM++, which provides more accurate localization through a
weighted combination of positive partial derivatives, making it particularly
suitable for medical imaging where precise localization matters.

Concept-based explanations offer an alternative approach by relating model
predictions to human-understandable concepts. Kim et al.
\cite{kim2018interpretability} introduced Testing with Concept Activation
Vectors (TCAV), which learns directions in a network's feature space
corresponding to user-defined concepts. This approach has been applied to
medical imaging to explain predictions in terms of clinically meaningful
attributes. Our FastCAV implementation adapts this framework for efficient
concept-based explanations aligned with dermatological concepts.

\subsection{Uncertainty Quantification in Deep Learning}

Reliable uncertainty estimates are essential for clinical decision support
systems, as they enable appropriate human-AI collaboration by identifying
cases requiring expert review. Gal and Ghahramani \cite{gal2016dropout}
demonstrated that dropout applied at test time approximates Bayesian
inference, providing uncertainty estimates without architectural changes.
This Monte Carlo Dropout approach has been widely adopted in medical imaging
due to its simplicity and effectiveness. Subsequent work has decomposed
uncertainty into epistemic (model uncertainty) and aleatoric (data
uncertainty) components \cite{kendall2017uncertainties}, providing more
nuanced characterization of prediction confidence. Our system implements
this decomposition to distinguish between cases where the model lacks
knowledge versus cases with inherently ambiguous features.

\section{Methods}

\subsection{Model Architecture}

MelanomaNet employs EfficientNet V2-M \cite{tan2021efficientnetv2} as the
backbone feature extractor. EfficientNet V2 improves upon the original
EfficientNet through training-aware neural architecture search and
progressive learning, achieving faster training and better parameter
efficiency. The medium variant provides 54 million parameters with
1280-dimensional feature outputs, balancing capacity with computational
requirements.

We process dermoscopic images at 384$\times$384 resolution, significantly
higher than the standard 224$\times$224 used in many classification tasks.
This higher resolution preserves fine details crucial for dermoscopic
analysis, such as subtle pigment patterns and border characteristics. The
classification head consists of global average pooling followed by dropout
(rate 0.3) and a linear layer mapping to 8 output classes corresponding to
ISIC 2019 categories: Melanoma (MEL), Nevus (NV), Basal Cell Carcinoma (BCC),
Actinic Keratosis (AK), Benign Keratosis (BKL), Dermatofibroma (DF), Vascular
lesion (VASC), and Squamous Cell Carcinoma (SCC). The dataset defines a ninth
Unknown (UNK) category, but no images are labeled with this class.

\begin{figure*}[t]
    \centering
    \includegraphics[width=0.95\textwidth]{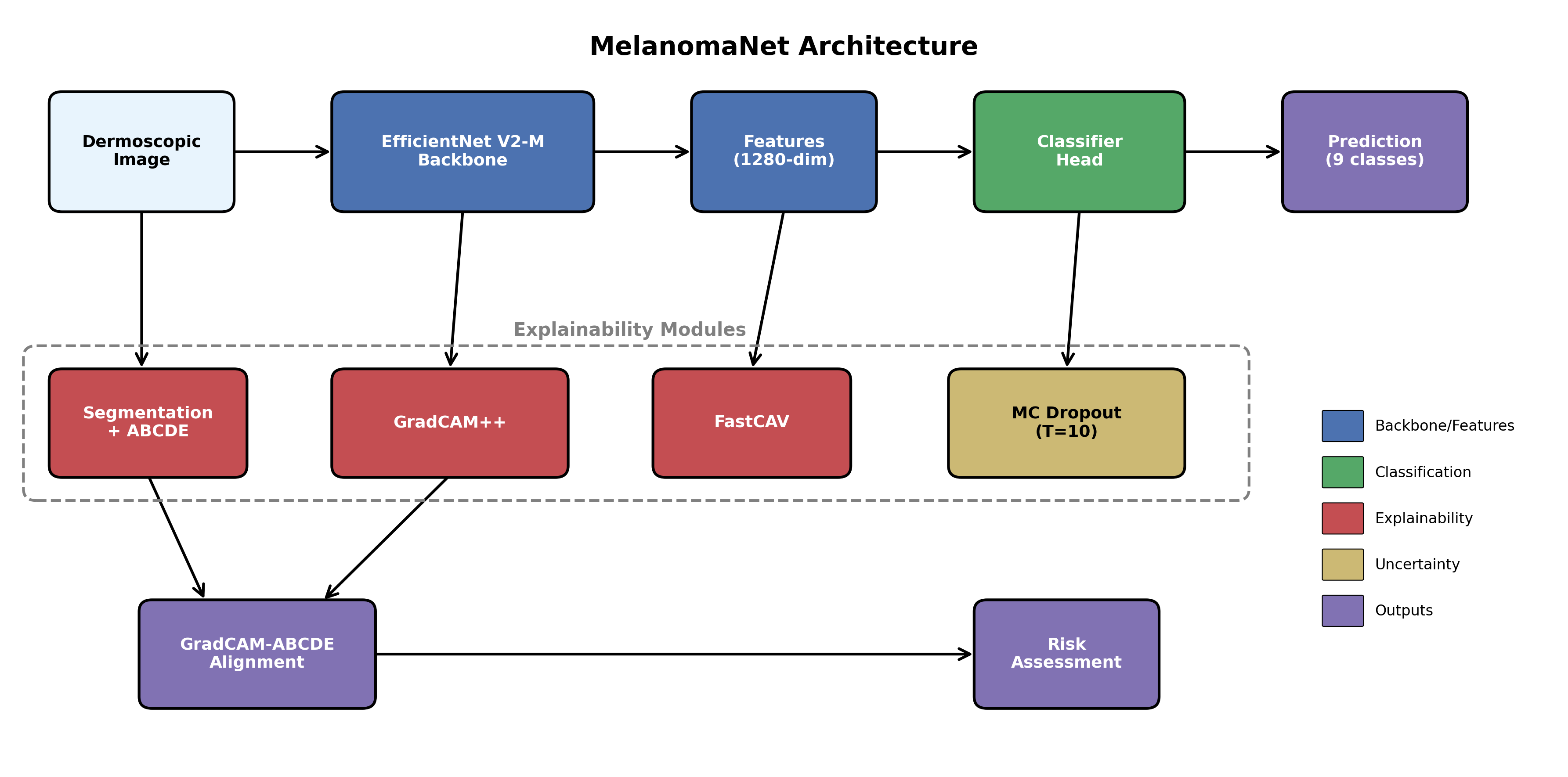}
    \caption{MelanomaNet system architecture. The main classification pipeline
        (top) processes dermoscopic images through EfficientNet V2-M to produce
        predictions across 8 classes. Four explainability modules (bottom, dashed box)
        provide complementary interpretations: ABCDE clinical criteria from image
        segmentation, GradCAM++ attention from feature maps, FastCAV concept scores
        from extracted features, and MC Dropout uncertainty estimates from stochastic
        forward passes. The GradCAM-ABCDE alignment module validates correspondence
        between model attention and clinical features.}
    \label{fig:architecture}
\end{figure*}

Figure \ref{fig:architecture} illustrates the complete system architecture,
showing how the classification pipeline integrates with four explainability
modules that operate on different levels of the model's representation.

\subsection{Training Configuration}

We train using weighted cross-entropy loss to address significant class
imbalance in the dataset, with class weights inversely proportional to class
frequencies. Optimization employs AdamW with initial learning rate $10^{-4}$,
weight decay $10^{-4}$, and cosine annealing schedule over 100 epochs. Mixed
precision training accelerates computation while maintaining numerical
stability. Data augmentation includes random horizontal and vertical flips,
rotation ($\pm20^\circ$), affine transformations (translation 10\%, scale
0.9--1.1), and color jittering (brightness, contrast, saturation 0.2, hue 0.1).

\subsection{GradCAM++ Attention Visualization}

We employ GradCAM++ \cite{chattopadhay2018grad} to visualize which image
regions most influence the model's predictions. GradCAM++ improves upon
GradCAM by computing pixel-wise importance weights for feature map
activations, providing better localization for multiple objects and sharper
attention maps. The method computes gradients of the predicted class score
with respect to the final convolutional layer's feature maps, then weights
and combines these activations to produce a class-discriminative heatmap.
The resulting visualization is upsampled to input resolution and overlaid
on the original image, allowing clinicians to verify that the model attends
to the lesion rather than spurious background features.

\subsection{ABCDE Clinical Criterion Analysis}

We implement automated extraction of ABCDE features to bridge model
predictions with clinical reasoning:

\textbf{Lesion Segmentation}: We extract lesion masks using Otsu's
thresholding with morphological refinement. Connected component analysis
removes small artifacts, and hole-filling produces robust binary masks.

\textbf{Asymmetry (A)}: We compare lesion halves along horizontal and
vertical axes through the centroid. The asymmetry score quantifies the
non-overlapping area between reflected halves, normalized to $[0,1]$.

\textbf{Border Irregularity (B)}: We analyze contour properties including
compactness (perimeter$^2$/area) and vertex count from polygon approximation.
Higher scores indicate more irregular borders.

\textbf{Color Variation (C)}: K-means clustering ($k=6$) identifies distinct
colors within the lesion. We count colors exceeding 5\% coverage and compute
color standard deviation as a variation score.

\textbf{Diameter (D)}: We compute the minimum enclosing circle and bounding
box diagonal, reporting the maximum extent in pixels.

\textbf{Risk Stratification}: A composite risk score sums binary flags from
each criterion (thresholds: asymmetry $>0.3$, border $>0.4$, colors $>3$,
diameter $>114$ pixels). Scores $\geq3$ indicate high risk, $\geq2$ medium
risk, and $<2$ low risk.

\subsection{GradCAM-ABCDE Alignment}

To validate that model attention corresponds to clinically relevant features,
we compute alignment metrics between GradCAM heatmaps and ABCDE analysis:

\begin{equation}
    \text{Alignment}_{\text{border}} = \frac{\sum_{i,j} M_{\text{border}}(i,j)
        \cdot H(i,j)}{\sum_{i,j} M_{\text{border}}(i,j)}
\end{equation}

where $M_{\text{border}}$ is a dilated border mask and $H$ is the normalized
attention heatmap. Similar metrics quantify attention overlap with the
overall lesion region.

\subsection{Fast Concept Activation Vectors}

We implement FastCAV for concept-based explanations, adapting TCAV
\cite{kim2018interpretability} with SGD classifiers for efficiency. For each
clinical concept (asymmetry, irregular border, multicolor, large diameter),
we train a linear classifier to distinguish concept-positive from
concept-negative examples in the model's 1280-dimensional feature space.
The Concept Activation Vector (CAV) is the normal to the learned decision
boundary.

The TCAV score quantifies concept influence by measuring the fraction of
inputs for which the model's prediction becomes more confident when moving
in the CAV direction. Positive scores indicate the concept supports the
prediction; negative scores indicate opposition. This provides
human-interpretable explanations showing which clinical attributes drive
each classification decision.

\subsection{MC Dropout Uncertainty Quantification}

We employ Monte Carlo Dropout \cite{gal2016dropout} with $T=10$ stochastic
forward passes to estimate uncertainty. By keeping dropout active during
inference, each forward pass samples from an approximate posterior
distribution over model weights, enabling Bayesian uncertainty estimation
without explicit probabilistic modeling. For each input, we compute three
complementary uncertainty measures:

\textbf{Predictive Uncertainty} captures total model uncertainty through
the entropy of averaged predictions:
\begin{equation}
    H[\bar{p}] = -\sum_c \bar{p}_c \log \bar{p}_c, \quad
    \bar{p} = \frac{1}{T}\sum_{t=1}^T p_t
\end{equation}
where $p_t$ is the softmax output from forward pass $t$, $\bar{p}$ is the
mean prediction across all $T$ passes, and $c$ indexes over classes. Higher
entropy indicates the model is uncertain about which class to predict.

\textbf{Epistemic Uncertainty} measures model uncertainty arising from
limited training data:
\begin{equation}
    \text{Var}[p] = \frac{1}{T}\sum_{t=1}^T (p_t - \bar{p})^2
\end{equation}
This variance across stochastic forward passes quantifies how much the
model's predictions fluctuate when different subsets of neurons are dropped.
High epistemic uncertainty suggests the model lacks sufficient training
examples similar to the input, and can be reduced with more data.

\textbf{Aleatoric Uncertainty} captures inherent data noise that cannot
be reduced:
\begin{equation}
    \frac{1}{T}\sum_{t=1}^T H[p_t]
\end{equation}
This averages the entropy of individual predictions, reflecting uncertainty
due to ambiguous or overlapping features in the input itself. Unlike
epistemic uncertainty, aleatoric uncertainty persists regardless of
training data quantity.

Predictions with predictive uncertainty exceeding threshold 0.5 are flagged
as unreliable, prompting clinical review. This decomposition allows
clinicians to understand whether uncertainty stems from model limitations
(epistemic) or inherent case ambiguity (aleatoric).

\section{Experiments and Results}

\subsection{Dataset}

We evaluate on the ISIC 2019 Challenge dataset
\cite{combalia2019bcn20000,tschandl2018ham10000} containing 25,331
dermoscopic images across 8 diagnostic categories. The class distribution
exhibits significant imbalance: Nevus (NV) comprises 50.83\% of samples
while Dermatofibroma (DF) represents only 0.94\%. We employ stratified
splitting with 70\% training (17,731 images), 15\% validation (3,800
images), and 15\% test (3,800 images).

\begin{table}[t]
    \centering
    \caption{Class distribution in the ISIC 2019 dataset.}
    \label{tab:distribution}
    \begin{tabular}{lrr}
        \toprule
        Class                         & Count  & Percentage \\
        \midrule
        NV (Nevus)                    & 12,875 & 50.83\%    \\
        MEL (Melanoma)                & 4,522  & 17.85\%    \\
        BCC (Basal Cell Carcinoma)    & 3,323  & 13.12\%    \\
        BKL (Benign Keratosis)        & 2,624  & 10.36\%    \\
        AK (Actinic Keratosis)        & 867    & 3.42\%     \\
        SCC (Squamous Cell Carcinoma) & 628    & 2.48\%     \\
        VASC (Vascular)               & 253    & 1.00\%     \\
        DF (Dermatofibroma)           & 239    & 0.94\%     \\
        \bottomrule
    \end{tabular}
\end{table}

\subsection{Classification Performance}

Table \ref{tab:results} presents overall classification metrics on the test
set. Our model achieves 85.61\% accuracy with weighted precision, recall,
and F1 scores all above 0.85, demonstrating strong performance despite
severe class imbalance.

\begin{table}[t]
    \centering
    \caption{Overall test set performance metrics.}
    \label{tab:results}
    \begin{tabular}{lc}
        \toprule
        Metric               & Value  \\
        \midrule
        Accuracy             & 0.8561 \\
        Precision (weighted) & 0.8600 \\
        Recall (weighted)    & 0.8561 \\
        F1 Score (weighted)  & 0.8564 \\
        \bottomrule
    \end{tabular}
\end{table}

Table \ref{tab:perclass} provides per-class breakdown. The model performs
best on Nevus (F1=0.91) and BCC (F1=0.89), which benefit from larger sample
sizes. Melanoma detection achieves F1=0.77 with 80.57\% precision and 74.52\%
recall. Minority classes show variable performance---DF achieves surprisingly
strong recall (86.11\%) despite having only 36 test samples, while AK
struggles with 63.85\% recall.

\begin{table}[t]
    \centering
    \caption{Per-class classification performance.}
    \label{tab:perclass}
    \begin{tabular}{lcccr}
        \toprule
        Class     & Precision & Recall & F1     & Support \\
        \midrule
        MEL       & 0.8057    & 0.7452 & 0.7743 & 679     \\
        NV        & 0.9255    & 0.9011 & 0.9131 & 1931    \\
        BCC       & 0.8579    & 0.9319 & 0.8934 & 499     \\
        AK        & 0.7545    & 0.6385 & 0.6917 & 130     \\
        BKL       & 0.6900    & 0.8270 & 0.7523 & 393     \\
        DF        & 0.6200    & 0.8611 & 0.7209 & 36      \\
        VASC      & 0.8947    & 0.8947 & 0.8947 & 38      \\
        SCC       & 0.8519    & 0.7340 & 0.7886 & 94      \\
        \midrule
        Macro avg & 0.8000    & 0.8167 & 0.8036 & 3800    \\
        \bottomrule
    \end{tabular}
\end{table}

\subsection{Explainability Outputs}

\begin{figure*}[t]
    \centering
    \includegraphics[width=0.95\textwidth]{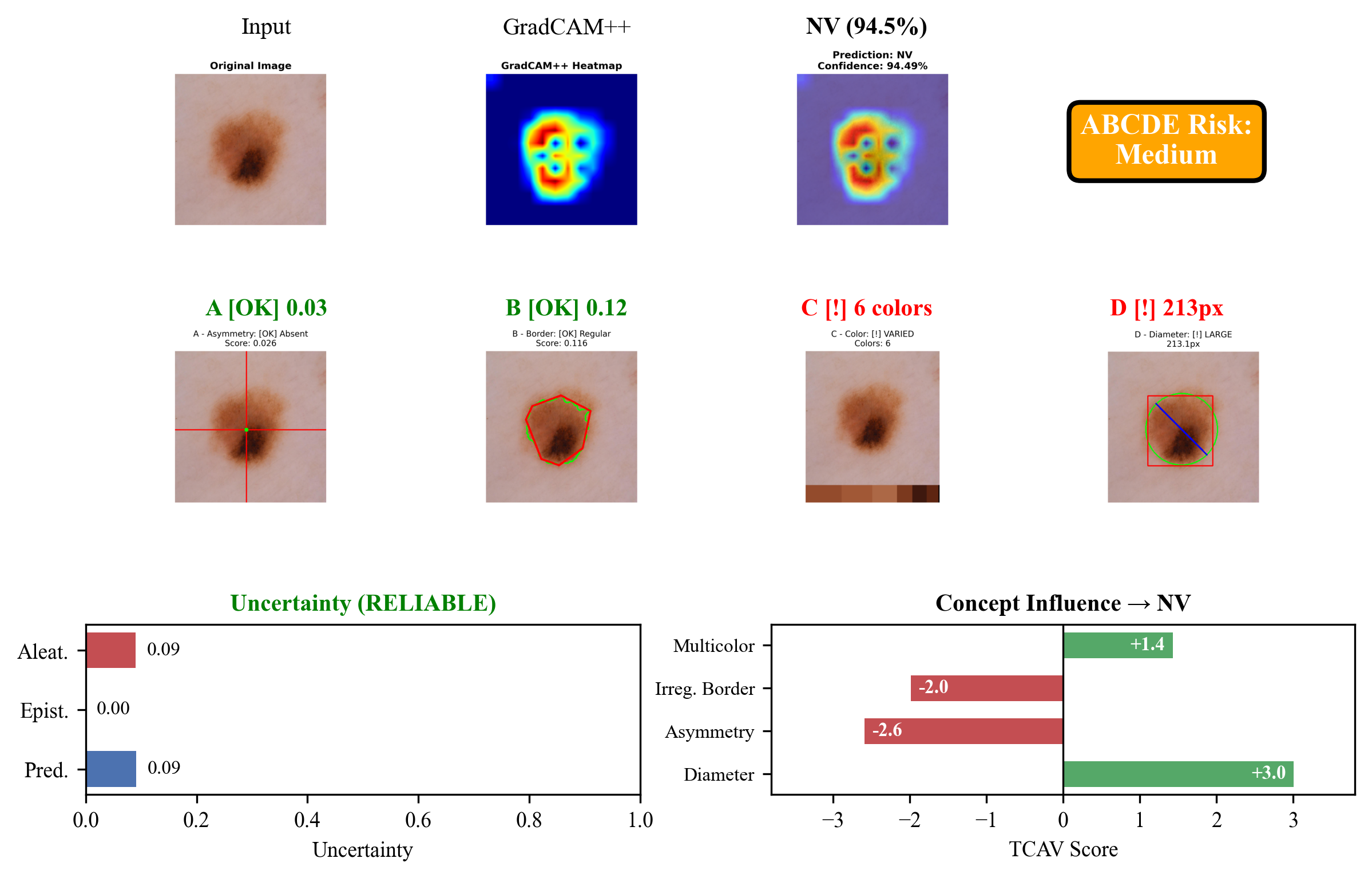}
    \caption{Analysis of a benign nevus (NV) with medium ABCDE risk. The model
        classifies with 94.49\% confidence. Top row: original image, GradCAM++
        heatmap, and overlay with prediction. Middle row: ABCDE criterion
        visualizations showing asymmetry axes, border contour, color palette, and
        diameter measurement. Bottom panels: uncertainty decomposition (left) and
        FastCAV concept importance scores (right).}
    \label{fig:sample_benign}
\end{figure*}

\begin{figure*}[t]
    \centering
    \includegraphics[width=0.95\textwidth]{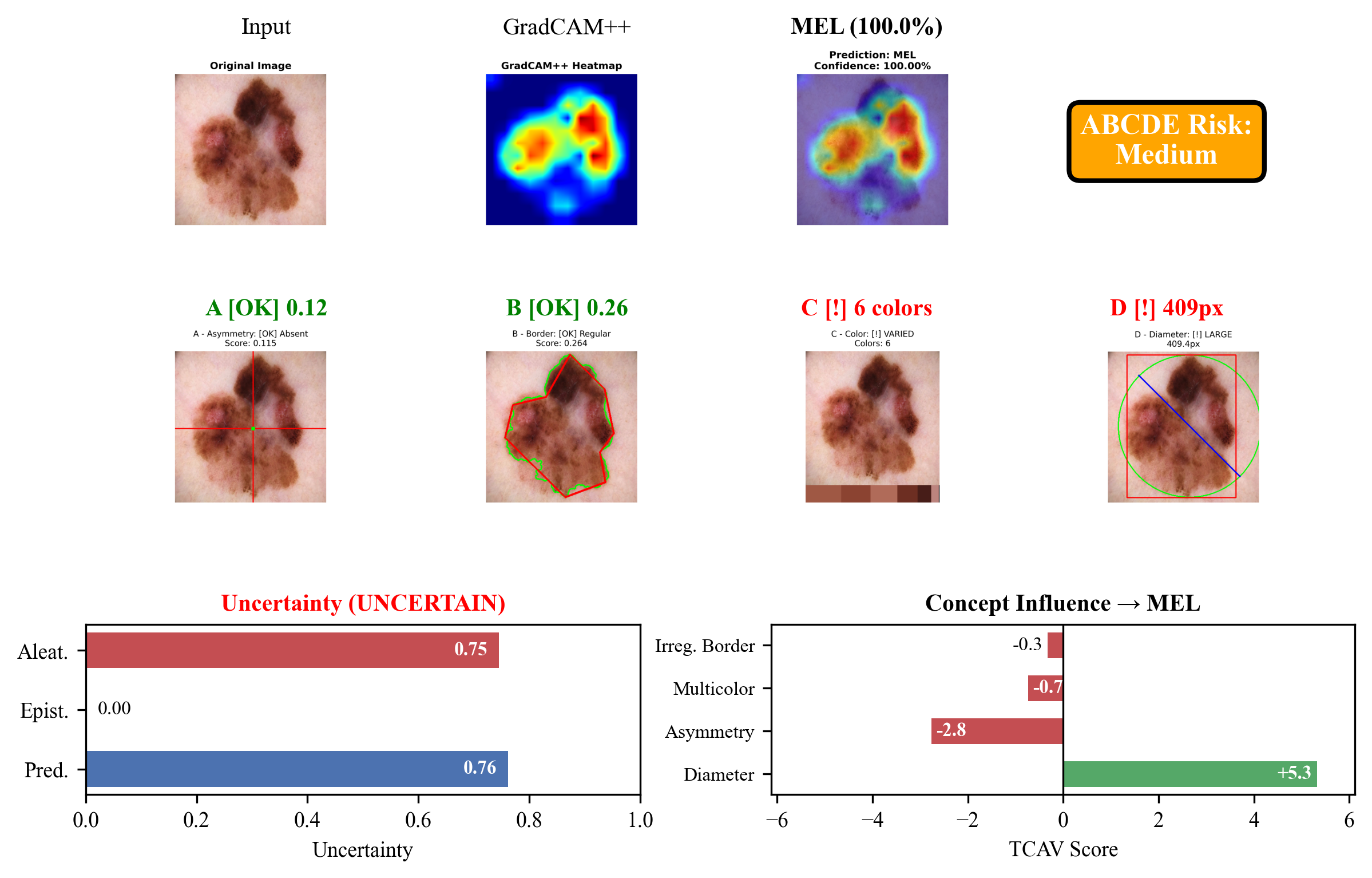}
    \caption{Analysis of a melanoma (MEL) with medium ABCDE risk. The model
        classifies with 100\% confidence but the uncertainty module flags this as
        ``UNCERTAIN'' due to high aleatoric uncertainty (0.75). ABCDE criteria show
        acceptable asymmetry (0.12) and borders (0.26), but flag multiple colors (6)
        and large diameter (409px). FastCAV analysis reveals large diameter strongly
        supports the prediction (+5.32) while asymmetry opposes it (-2.77).}
    \label{fig:sample_mel}
\end{figure*}

Figures \ref{fig:sample_benign} and \ref{fig:sample_mel} demonstrate the
comprehensive analysis output for test images with different risk profiles
and reliability assessments.

Figure \ref{fig:sample_benign} shows a correctly classified nevus with
94.49\% confidence. The GradCAM++ heatmap confirms the model focuses on the
lesion center. ABCDE analysis shows low asymmetry (0.026), regular borders
(0.116), but multiple colors (6) and larger diameter (213 pixels), yielding
medium risk. Uncertainty analysis reports low predictive uncertainty (0.088),
flagged as ``RELIABLE.'' FastCAV scores show large diameter (+2.29) and
multicolor (+0.53) supporting the prediction while asymmetry (-1.43) and
irregular border (-1.31) oppose it.

Figure \ref{fig:sample_mel} presents a melanoma classified with 100\%
confidence. Despite high confidence, the uncertainty module flags this as
``UNCERTAIN'' due to high predictive uncertainty (0.76) dominated by
aleatoric uncertainty (0.75), demonstrating that confidence and uncertainty
provide complementary information. ABCDE analysis shows acceptable asymmetry
(0.12) and regular borders (0.26), but flags multiple colors (6) and large
diameter (409 pixels). FastCAV concept scores show large diameter strongly
supports the melanoma prediction (+5.32) while asymmetry (-2.77), multicolor
(-0.74), and irregular border (-0.33) oppose it.

\subsection{Alignment Validation}

The GradCAM-ABCDE alignment metrics validate that model attention corresponds
to clinically relevant features. Across test samples, we observe mean lesion
attention of 0.60, indicating the model predominantly focuses within the
segmented lesion region rather than background. Border alignment scores
(mean 0.53) suggest moderate attention to lesion boundaries, consistent
with border features contributing to classification.

\section{Discussion and Summary}

We presented MelanomaNet, an explainable deep learning system for multi-class
skin lesion classification that combines strong classification performance
with comprehensive interpretability. Our approach addresses a key barrier to
clinical adoption by providing multiple complementary explanation modalities
that relate model predictions to familiar clinical concepts.

\textbf{Clinical Relevance}: The ABCDE criterion analysis bridges the gap
between deep learning predictions and established dermatological assessment.
By automatically extracting and quantifying these clinical features, our
system generates explanations that clinicians can evaluate using their
domain expertise. The alignment metrics provide validation that the model
attends to clinically meaningful image regions.

\textbf{Uncertainty Awareness}: The MC Dropout uncertainty quantification
enables appropriate human-AI collaboration by identifying predictions that
warrant expert review. The epistemic/aleatoric decomposition offers
additional insight---high epistemic uncertainty suggests cases outside the
training distribution, while high aleatoric uncertainty indicates inherently
ambiguous images requiring careful clinical evaluation.

\textbf{Concept-Based Reasoning}: FastCAV concept scores provide an intuitive
explanation format, indicating which clinical concepts support or oppose each
prediction. This approach complements attention visualization by
characterizing predictions in semantic terms rather than spatial regions.

\textbf{Limitations}: Several limitations merit discussion. The ABCDE
analysis relies on automated segmentation, which can fail for lesions with
low contrast or complex backgrounds. Evolution (the E criterion) requires
temporal image sequences unavailable in single-image datasets. Class
imbalance remains challenging---minority classes like AK achieve lower
recall despite weighted loss. Finally, while our explainability mechanisms
provide useful insights, they cannot guarantee that the model's internal
reasoning truly follows clinical logic.

\textbf{Future Work}: Potential extensions include incorporating temporal
analysis for evolution tracking, expanding the concept vocabulary for
FastCAV, and conducting user studies with dermatologists to evaluate
clinical utility.

In conclusion, MelanomaNet demonstrates that high classification accuracy
and comprehensive interpretability are not mutually exclusive. By providing
GradCAM++ attention visualization, ABCDE clinical criterion analysis,
concept-based explanations, and uncertainty quantification, our system
offers a multi-faceted approach to explainable dermoscopic image analysis
that could facilitate greater trust and adoption in clinical workflows.

{\small
\bibliographystyle{ieeetr}
\bibliography{references}
}

\end{document}